%% file: zazo.tex
\documentclass{article}

\include{preamble}

\title{CONVOLUTIONAL DICTIONARY LEARNING IN HIERARCHICAL NETWORKS}
\name{Javier Zazo, \qquad Bahareh Tolooshams, \qquad Demba Ba} 
\address{Harvard John A. Paulson\\School of Engineering and Applied Sciences\\Harvard University}
\begin{document}
%
\maketitle
\begin{abstract}
Filter banks are a popular tool for the analysis of piecewise smooth signals such as natural images.
Motivated by the empirically observed properties of scale and detail coefficients of images in the wavelet domain, we propose a hierarchical deep generative model of piecewise smooth signals that is a recursion across scales: the low pass scale coefficients at one layer are obtained by filtering the scale coefficients at the next layer, and adding a high pass detail innovation obtained by filtering a sparse vector.
This recursion describes a linear dynamic system that is a non-Gaussian Markov process across scales and is closely related to multilayer-convolutional sparse coding (ML-CSC) generative model for deep networks, except that our model allows for deeper architectures, and combines sparse and non-sparse signal representations.
We propose an alternating minimization algorithm for learning the filters in this hierarchical model given observations at layer zero, e.g., natural images.
The algorithm alternates between a coefficient-estimation step and a filter update step. The coefficient update step performs sparse (detail) and smooth (scale) coding and, when unfolded, leads to a deep neural network.
We use MNIST to demonstrate the representation capabilities of the model, and its derived features (coefficients) for classification.
\end{abstract}
\begin{keywords}
Convolutional dictionary learning, sparse coding, deep networks, hierarchical models.
\end{keywords}
\section{Introduction}
\label{sec:intro}

With the advent of neural networks and current state-of-the-art performance on many machine learning applications \cite{LeCun2015}, deep learning has become an ubiquitous framework with which to address problems in a wide range of domains.
In particular, convolutional neural networks (CNNs) have been very successful for image classification \cite{Krizhevsky2012}, 
as they are able to reduce the number of trainable parameters, and still capture latent representations for discriminative problems.

However, little is known about how to obtain more efficient representations, or on how to train smaller networks that perform as good as CNNs, and not require exhaustive architecture search \cite{Frankle2018}.
The importance of such advancement lies not only on obtaining more systematic, interpretable and efficient models, but also on reducing the economic and ecological footprint of neural networks \cite{Strubell2019}. 

Representation analysis with wavelets is a classical, interpretable and well understood theory that allows to decompose images/signals into a linear combination of basis functions at different scales to represent an image, and recover it perfectly via convolution operations \cite{Mallat1989ATF,Mallat1999}.

Alternatively, convolutional sparse coding (CSC) \cite{BristowHilton2013FCSC} and convolutional dictionary learning (CDL) \cite{garcia-2018-convolutional} use sparse representation of images and learned dictionaries from the images of the database.
More recently, \cite{SulamJeremias2018MCSM} uses sparse representations and dictionaries obtained at different levels of a network. 

This paper proposes a convolutional generative hierarchical model of signals, e.g., images, where the filters are learned from the data, and the images are decomposed into \emph{scale} and \emph{detail} signals.
Additionally, there is a one-to-one correspondence between the sparse coding, or inference step, and deep CNNs.
Such representation is inspired by a combination of wavelet analysis \cite{Mallat1999}, sparse coding \cite{JCandes2006}, and dictionary learning \cite{agarwal2013exact}.
The \emph{scale} consists of a dense signal, while the \emph{detail} is a sparse representation that selects a few dictionary atoms.

\noindent\textbf{Related work:}
A notable precursor to our work are \emph{deconvolutional networks} \cite{Zeiler2010DeconvolutionalN}, which described a generative hierarchical model inspired by CNNs where, starting with a sparse signal encoding, an image is obtained through the cascade of convolutional operations.
However, a significant limitation of such networks is that the attainable levels of sparsity at each layer is reduced with the depth of the architecture.

The formal analysis of CSC is introduced by \cite{Papyan2017}, which developed sufficient theoretical guarantees for exact sparse signal recovery.
The sparse coding problem is non-convex and NP-hard \cite{Natarajan1995}, but when specific sparsity levels of the signal are satisfied (w.r.t. to the mutual coherence of the employed dictionaries) an optimal solution can be retrieved \cite{JCandes2006}.
In such case, a relaxed formulation of the problem yields optimal results, making it possible to recover the generating signal.
However, the feature maps (signal encodings) from stacks of convolutional layers become less sparse as the model becomes deeper, which renders solving the sparse inverse problem harder with increasing depth.

The work by Sulam et. al \cite{SulamJeremias2018MCSM} addresses this problem of multilayer CSC (ML-CSC) by enforcing sparsity also across dictionaries. 
If the convolutional filters are very sparse, the subsequent layers will contain a reduced number of non-zeros, and some guarantees for a unique representation can be established.
However, this model requires above 99\% dictionary sparsity levels, and a large number of channels.

Our proposed model builds on top of existing results by considering an inverse problem that decomposes a source image into smooth and sparse signals jointly.
As previously mentioned, this decomposition is quite natural in wavelet analysis, or scattering networks \cite{Bruna2013}, and combines the tools of CSC and CDL together.
We remark that the deconvolutional model \cite{Zeiler2010DeconvolutionalN}, and the ML-CSC model \cite{SulamJeremias2018MCSM} cannot have arbitrary depths, because of limitations on the signal sparsity levels.
In contrast, our proposed model is not limited by depth as sparse details are added separately at every layer, and the recursion is on the scale, which does not have to be sparse.

%
%

\section{Model Description}
\label{sec:model}

Given a \emph{scale} signal $\bx_{L}$ and \emph{detail} signals $\bu =[ \bu_1,\dots, \bu_{L}]$, we propose the following recursive generative model
\begin{equation}\label{eq:recursive-model}
	\bx_{\ell-1} = \bA_{\ell} \ast \bx_{\ell} + \bB_{\ell} \ast \bu_{\ell} + \bEps_{\ell}, \quad \ell\in\set{1,\ldots, L},
\end{equation}
and assume the following latent prior distributions:
\begin{IEEEeqnarray}{rCl}\IEEEyesnumber \label{eq:priors}
	\bEps_{\ell} & \sim & \calN(0,\sigma_{\ell}^2), \qquad \forall \ell \in\set{1,\ldots, L} \IEEEyessubnumber  \\
	\bu_{\ell} & \sim & \laplace(0, \lambda_{\ell}),
	\IEEEyessubnumber \\
	\bx_{\ell} & \sim & \calN(0, \sigma_{x_{\ell}}^2).\IEEEyessubnumber 
\end{IEEEeqnarray}
Here, $ \ell $ indicates layer index, where $ \ell=0 $ refers to the input signal, and $ \ell>0 $ refers to a deeper encoding. 
The model has total depth $ L $, and $ \ast $ indicates the full convolution operation between two signals.
We remark that the model given by Eq.~\eqref{eq:recursive-model} is a non-Gaussian Markovian dynamical system~\cite{Dethlefsen2003}.

We represent filters with capital bold letters, i.e., $ \bA_{\ell} $ and $ \bB_{\ell} $, and signals with lower case bold letters, i.e., $ \bx_{\ell} $, $ \bu_{\ell} $.
Filters are tensors of dimensions $C_{\ell}\times D_{\ell}\times H_{\ell}\times W_{\ell} $, 
referring to the number of output channels, depth (input channels), height and width of filters, respectively.
Explicitly, the convolution operation involves the following computation,
\begin{equation}\label{key}
	(\bA_{\ell} \ast \bx_{\ell})(r) = \sum_{c=1}^{D_{\ell}} \bA_{\ell}(r,c) \ast \bx_{\ell}(c), \quad \forall r\in {1,\ldots,C_{\ell}},
\end{equation}
where $ c $ indexes the input feature map channels, and $ r $ indicates output channels.
We simplify the whole convolution operation by simply writing $ \bA_{\ell} \ast \bx_{\ell} $. 

Eq. \eqref{eq:recursive-model} indicates a recursive relation between input and output signals across layers.
We refer $ \bx_{\ell} $ ($ \ell\geq 1 $) the \emph{scale} signal at layer $ \ell$.
It contributes smoothly to $ \bx_{\ell-1} $ because of the averaging of convolution and its non-sparse form. 

Eq. \eqref{eq:priors} further specifies priors for our model.
For example, $ \bu_{\ell} $ is assumed to follow a Laplace distribution of mean zero and \emph{diversity} coefficient $\lambda_{\ell} $.
This prior supports that $ \bu_{\ell} $ is sparse, and adds high frequency information when convolved with $ \bB_{\ell} $ to the signal $ \bx_{\ell} $.
We refer to $ \bu_{\ell} $ as \emph{detail} signal, following wavelet terminology.

\noindent\textbf{Hierachical model with tied filters}: A simple modification can be made on Eq. \eqref{eq:recursive-model} to resemble wavelet analysis, by tying filters $ \bA_{\ell} = \bA_{\ell+1} $ and $ \bB_{\ell} = \bB_{\ell+1} $ across layers for $ \ell\in\set{1,\ldots,L-1} $, similar to a wavelet being repeatedly used across decomposition levels.
Additionally, up/down sampling operations can be incorporated to obtain a multiscale CSC/CDL model, although such analysis is out of the scope of this paper.

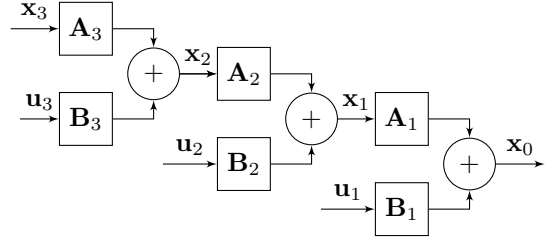
\begin{figure}
	\centering
	\input{model.tikz}
	\caption{Hierarchical model representation for $ L=3 $.}
	\label{fig:hierarchical}
	\vspace{-1em}
\end{figure}

\Cref{fig:hierarchical} summarizes the model described by Eqs. \eqref{eq:recursive-model} and \eqref{eq:priors} for three layers.
As we noted in the related work (\cref{sec:intro}), our model does not have limitations in terms of attainable depth restricted by sparsity requirements.
This is because all $ \bu_{\ell} $ terms are independent across layers.

Finally, we note that by setting $ \bB_{\ell} = 0$ for all layers in Eq.~\eqref{eq:recursive-model} and establishing a prior such that $ \bx_{\ell} \sim\laplace(0,\lambda_{\ell}) $ ($ \ell\geq 1 $), our model simplifies to a deconvolutional network \cite{Zeiler2010DeconvolutionalN}.
By further imposing sparsity on the filters $ \bA_{\ell} $, then our model simplifies to ML-CSC \cite{SulamJeremias2018MCSM}.

\section{Hierarchical CSC}
\label{sec:csc}

We can synthetize images from scale representation $ \bx_L $ and detail signals across layers $ \bu \triangleq [\bu_1,\ldots,\bu_L] $ using Eq. \eqref{eq:recursive-model}.
However, the analysis step requires solving the inverse problem to find appropriate encodings for an image $ \bx_0 $ across layers, i.e., $ \bx \triangleq [\bx_1, \ldots, \bx_L]$ and $ \bu $.
We refer to such problem as hierarchical convolutional sparse coding (H-CSC).

The correct representation for scale and detail signals can be obtained by maximizing the log-posterior of the state-sequence $ \bx $ and the input $ \bu $ from our model (assuming fixed filters $ \bA_{\ell}$, $ \bB_{\ell}$).
The problem is coupled by the scale signals in $ \bx $, but for simplicity we solve each layer separately, similar to \cite{SulamJeremias2018MCSM,Zeiler2010DeconvolutionalN}.
This leads to the following problem  
\begin{equation}\label{eq:analysis}
	\hat{\bx}_{\ell},\hat{\bu}_{\ell} = \argmin_{\bx_{\ell}, \bu_{\ell}} \: f(\hat{\bx}_{\ell-1} ,\bx_{\ell},\bu_{\ell}) + \lambda_{\ell} \Vert \bu_{\ell} \Vert_1 + {\gamma_{\ell}} \Vert \bx_{\ell} \Vert^2_2,
\end{equation}
\begin{equation}\label{eq:smooth-f}
	f(\hat{\bx}_{\ell-1} , \bx_{\ell},\bu_{\ell}) = \frac{1}{2}\Vert \hat{\bx}_{\ell-1} - \bA_{\ell} \ast \bx_{\ell} - \bB_{\ell} \ast \bu_{\ell} \Vert^2_2.
\end{equation}
for every layer $ \ell\in\set{1,\ldots,L} $.
Here, $ \hat{\bx}_0=\bx_0 $ is given as input image, and subsequent estimates $ \hat{\bx}_{\ell} $ are obtained after solving Eq. \eqref{eq:analysis}.

\noindent\textbf{Relationship with CNNs and ReLU activations}: Eq. \eqref{eq:analysis} incorporates two important regularizers into the model.
The $ \ell_1 $-norm enforces sparsity on $ \bu_{\ell} $ signal, and accomplishes this result with high resemblance to standard CNNs.
Consider the solution to the following problem,
\begin{equation}\label{eq:relu-sol}
	\min_{\bu} \quad \frac{1}{2}\Vert \bu - \bb\Vert^2_2 + \lambda\Vert \bu \Vert_1,
\end{equation}
which can be written succinctly via soft-thresholding:
\begin{IEEEeqnarray}{rCl}\label{eq:soft-thresholding}
	\mathcal{S}_{\lambda}(\bb)&=& 
	\relu(\bb-\lambda)-\relu(-\bb-\lambda).
\end{IEEEeqnarray}
The equivalence of the soft-thresholding operation, emphasizes that CNNs with ReLU activations have similar response as regularized problems like Eq. \eqref{eq:analysis}.
This remark has been previously discussed in \cite{PapyanV2017CNNA,Ba2018}, and justifies our motivation to induce a sparse prior on $ \bu_{\ell} $.
A second regularizer $\gamma_{\ell} \Vert \bx_{\ell+1} \Vert^2_2$ conveys smoothness, but also guarantees uniqueness of the solution if the problem is not strongly convex.

\noindent\textbf{FISTA derivation:} Because Eq. \eqref{eq:analysis} is non-smooth and convolutional, it can be solved via iterative proximal algorithms.
These methods evaluate the gradient on the smooth part of the function, and apply a proximal operation on the non-smooth part (such as soft-thresholding).
Accelerated algorithms like FISTA \cite{beck2009fast} incorporate past estimates in the update formula, and achieve faster convergence speeds.

The FISTA algorithm admits an efficient implementation on GPUs with known gradients, and can parallelize the computations across examples.
This is particularly useful because GPUs implement convolution operations efficiently, and we can exploit these subroutines to reduce the computational requirements.
The whole procedure is detailed in \cref{alg:fista}, where $ \alpha $ denotes an appropriate step-size.
The algorithmic derivation requires obtaining the gradients on Eq. \eqref{eq:smooth-f} w.r.t. $ \bu_{\ell} $ and $ \bx_{\ell} $.
Specifically, we get 
\begin{equation}\label{eq:gradient-ul}
\nabla_{\bu_{\ell}} f(\hat{\bx}_{\ell-1},\bx_{\ell},\bu_{\ell}) = (\bA_{\ell} \ast \bx_{\ell} + \bB_{\ell} \ast \bu_{\ell} - \bx_{\ell-1} ) \star \bB_{\ell},
\end{equation}
\begin{equation}\label{eq:gradient-xl}
\nabla_{\bx_{\ell}} f(\hat{\bx}_{\ell-1},\bx_{\ell},\bu_{\ell}) = (\bA_{\ell} \ast \bx_{\ell} + \bB_{\ell} \ast \bu_{\ell} - \bx_{\ell-1} ) \star \bA_{\ell}.
\end{equation}
where $ \star $ refers to \emph{valid} correlation between two signals.


\begin{algorithm}[!tb]
	\caption{FISTA algorithm for H-CSC \eqref{eq:analysis}.}
	\label{alg:fista}
	Input: $\bx_{\ell}$, $\alpha$, $ \lambda_{\ell} $, $ \gamma_{\ell} $ . \\
	Initialize: $\bx^1_{\ell+1}$, $\bu^1_{\ell+1}$, $ t^1=1 $. \\
	\For{$k \in \{1,\dots, K\}$}{
	$ t^{k+1} \leftarrow 1 + \sqrt{1+4(t^{k})^2} / 2$ \\
	$ \overline{\bu} \leftarrow \bu^k_{\ell} + (t^k-1) (\bu^k_{\ell}-\bu^{k-1}_{\ell}) / t^{k+1}$ \\
	$ \overline{\bx} \leftarrow \bx^k_{\ell} + (t^k-1) (\bx^k_{\ell}-\bx^{k-1}_{\ell}) / t^{k+1}$ \\
	$ \br \leftarrow \bA_{\ell} \ast \overline{\bx} + \bB_{\ell} \ast \overline{\bu} -\bx_{\ell} $ \\
	$ \bx^{k+1}_{\ell} \leftarrow \overline{\bx} - \alpha (\br \star \bA_{\ell} + \gamma_{\ell} \overline{\bx}) $ \\
	$ \bu^{k+1}_{\ell} \leftarrow \mathcal{S}_{\lambda_{\ell} \alpha}(\overline{\bu} - \alpha (\br \star \bB_{\ell})) $
	}
	\textbf{return} $ \bx_{\ell}^K $, $ \bu_{\ell}^K $.
\end{algorithm}

\section{Convolutional Dictionary Learning}
\label{sec:cdl}

The negative of the log-posterior that results from the generative model presented by Eqs. \eqref{eq:recursive-model} and \eqref{eq:priors} is non-convex (bilinear) on both filters $ \bA_{\ell} $, $ \bB_{\ell} $ and variables $ \bx_{\ell} $, $ \bu_{\ell} $, across layers.
However, it is natural to propose an alternating optimization scheme that solves the problem on specific variables while fixing the rest of them.
We already described in \cref{sec:csc} how to solve on variables $ \bx_{\ell} $ and $ \bu_{\ell} $ for fixed filters $ \bA_{\ell} $ and $ \bB_{\ell} $, which we referred as the \emph{analysis} step.

To update the filter variables, a simple approach consists of fixing the scale and detail signals and updating the filters with a first-order gradient method.
The loss function is a concatenation of example images solving Eq. \eqref{eq:analysis}.
With a slight abuse of notation, we denote $ \hat{\bx}_{\ell,i} $, $ \hat{\bu}_{\ell,i} $ the $ i $'th encoding estimate across a database $\set{1,\ldots,N} $:
\begin{equation}\label{eq:cdl}
\begin{IEEEeqnarraybox}[][c]{r'l}
	\min_{\bA_{\ell},\bB_{\ell}} & \sum_{i=1}^N \Vert \hat{\bx}_{l-1,i} - \bA_{\ell} \ast \hat{\bx}_{l,i} - \bB_{\ell} \ast \hat{\bu}_{l,i} \Vert^2_2 \\
	\text{s.t.} & \bA_{\ell} \star \bA_{\ell} = 1,\quad \bB_{\ell} \star \bB_{\ell} = 1
\end{IEEEeqnarraybox}
\end{equation}

Updating the filters $ \bA_{\ell} $ and $ \bB_{\ell} $ requires computing gradients from Eq. \eqref{eq:cdl}.
Current approaches can exploit the autoencoder relation of a generative model (first finding a latent representation, then reconstructing) to obtain gradients through backpropagation and automatic differentiation \cite{TolooshamsBahareh2018SCDL}.
This autoencoder formulation allows to use GPUs directly with appropriate automatic differentiation software.

A second approach computes gradients in the Fourier domain, updates the filters, and converts the updated filters back to time domain via inverse Fourier transform \cite{garcia-2018-convolutional}.
To the best of our knowledge, this procedure does not currently run efficiently on GPU, making it inappropriate to use for large datasets or images.

\noindent\textbf{Filter gradients:} 
Our approach computes the gradients directly on the loss function Eq. \eqref{eq:cdl}.
This mechanism avoids automatic differentiation, which uses GPU memory and computation time, and also avoids converting from Fourier and back on every update.
Filters are four dimensional tensors $(C_{\ell}\times D_{\ell}\times H_{\ell}\times W_{\ell})$, and images three dimensional $(D_{\ell}\times H_{\ell}\times W_{\ell})$.
To compute the gradient w.r.t. the filter, we can extend with an extra dimension the image, and operate on 3D correlation (rather than 2D).
Then, we obtain the exact gradient expression as follows:
\begin{equation}\label{eq:filter-time-derivative}
\begin{IEEEeqnarraybox}[][c]{l}
	\br_{\ell,i} = ( \bA_{\ell} \ast \bx_{\ell,i} + \bB_{\ell} \ast \bu_{\ell,i} - \bx_{\ell-1,i}) \\ \frac{\partial}{\partial \bA_{\ell}} \Vert \bx_{\ell-1,i} - \bA_{\ell} \ast \bx_{l,i} - \bB_{\ell} \ast \bu_{\ell,i} \Vert^2_2 = \overline{\br}_{\ell,i} \star \overline{\bx}_{\ell,i}.
\end{IEEEeqnarraybox}
\end{equation}
Here, $ \overline{\br}_{\ell,i} $, $ \overline{\bx}_{\ell,i} $ denotes each of the variables with extra dimensions.
The previous expression provides an efficient way to compute gradients directly on GPU and update filters accordingly.
The gradient w.r.t. $ \bB_{\ell} $ has an analog form as Eq.~\eqref{eq:filter-time-derivative}.

\section{Experimental Results}
\label{sec:simulations}
\vspace{-0.5em}
To illustrate our results, we trained a set of hierarchical models with $L=3$ on MNIST database, comprising 60,000 training and 10,000 test grayscale digit images of $ 28\times 28 $ pixels.
The training step uses \cref{alg:fista} to solve the inverse problem, and then  updates the filters with stochastic gradient descent following \cref{sec:cdl}.
The whole filter training procedure is unsupervised, and minimizes the reconstruction error of the input images.

After the model had converged, we used H-CSC features to train a multiclass logistic regression classifier, using $ \bu $ and $ \bx $ as inputs.
Our classification results are shown in \cref{tab:mnist-accuracy}, reaching accuracies above 98.1\% on the test set with a 3 layer hierarchical model and tied filters.
We note that our classification results are similar with those reported by ML-CSC~\cite{SulamJeremias2018MCSM}.

We also indicate the number of trainable parameters on each model in \cref{tab:mnist-accuracy}.
The multiscale model has significantly less number of trainable parameters because the filters are repeated between layers.
On the other hand, ML-CSC used a 3 layer network with 1,664,800 trainable parameters, although most of them are zero.
What we show is that our model is capable of finding appropriate encodings with a recursive structure, tied filters and reduced number of parameters.

\noindent\textbf{Simulation parameters:}
Parameters of the model and algorithm were chosen with grid search, for a total of 12 simulated models.
The same paremeters were used in all layers.
The FISTA regularization $ \lambda $ was varied between 1.0 and $10^{-3}$, and $  \lambda_{\ell} =1.0 $ was selected as giving the best accuracy.
This result indicates that high sparsity levels help obtain better classification performance.

The FISTA learning rate $ \alpha $ was chosen between 0.01 and 0.001, and $ \alpha=0.01 $ yielded best results.
Similarly, we unfolded the network for $ K=\set{40,100} $ FISTA iterations at every layer for every dictionary gradient update, where the best parameter was $K=40$.
A larger number of FISTA iterations did not help achieve better accuracy results.

Finally, the scale filter was selected with $ 5\times 5 $ spatial dimensions, and single input and output channels.
The single channel was fixed to a constant value during training.
This construction aimed to detect the low frequencies of the images. 
On the other hand, the detail filter $ \bB_{\ell} $ had same spatial dimensions $ 5\times 5 $ and 32 output channels.
We found experimentally that reducing the number of scale filters improved the classification performance.
Still, having at least a single scale filter allows our model to reach arbitrary depths.

Our implementations are build on PyTorch and run on 1080-GTX or Titan XP Nvidia GPUs.



\begin{table}
	\centering
	\caption{Classification accuracy on MNIST.}
	\label{tab:mnist-accuracy}
	\begin{tabular}{|c|c|c|c|}
		\hline 
		Network models & Train Set & Test Set & Parameters\tabularnewline
		\hline 
		\hline 
		1 layer & 98.41 & 97.47 & 800\tabularnewline
		\hline 
		3 layers (tied) & 99.10 & 98.11 & 800\tabularnewline
		\hline 
		3 layers (ML-CSC) &  & 98.85 & 1,664,800\tabularnewline
		\hline 
	\end{tabular}
\vspace{-1em}
\end{table}

\begin{figure}[h!]
	\centering
	\begin{subfigure}[t]{.49\linewidth}
	\centering
	\includegraphics[width=1\columnwidth,height=0.7\columnwidth]{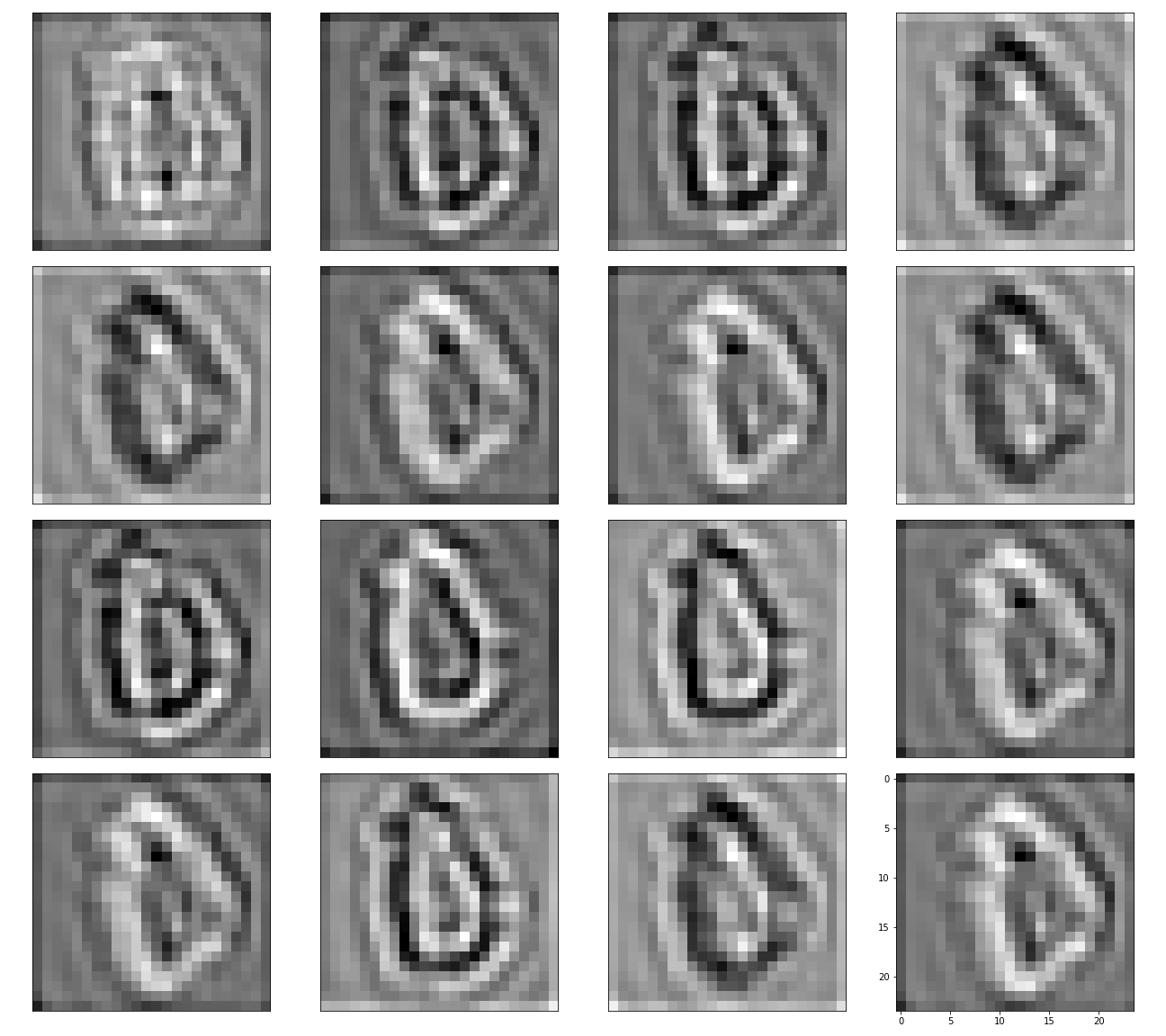}
	\caption{First layer $ \bx_1 $.}
	\end{subfigure}
	\begin{subfigure}[t]{.49\linewidth}
	\centering
	\includegraphics[width=1\columnwidth,height=0.7\columnwidth]{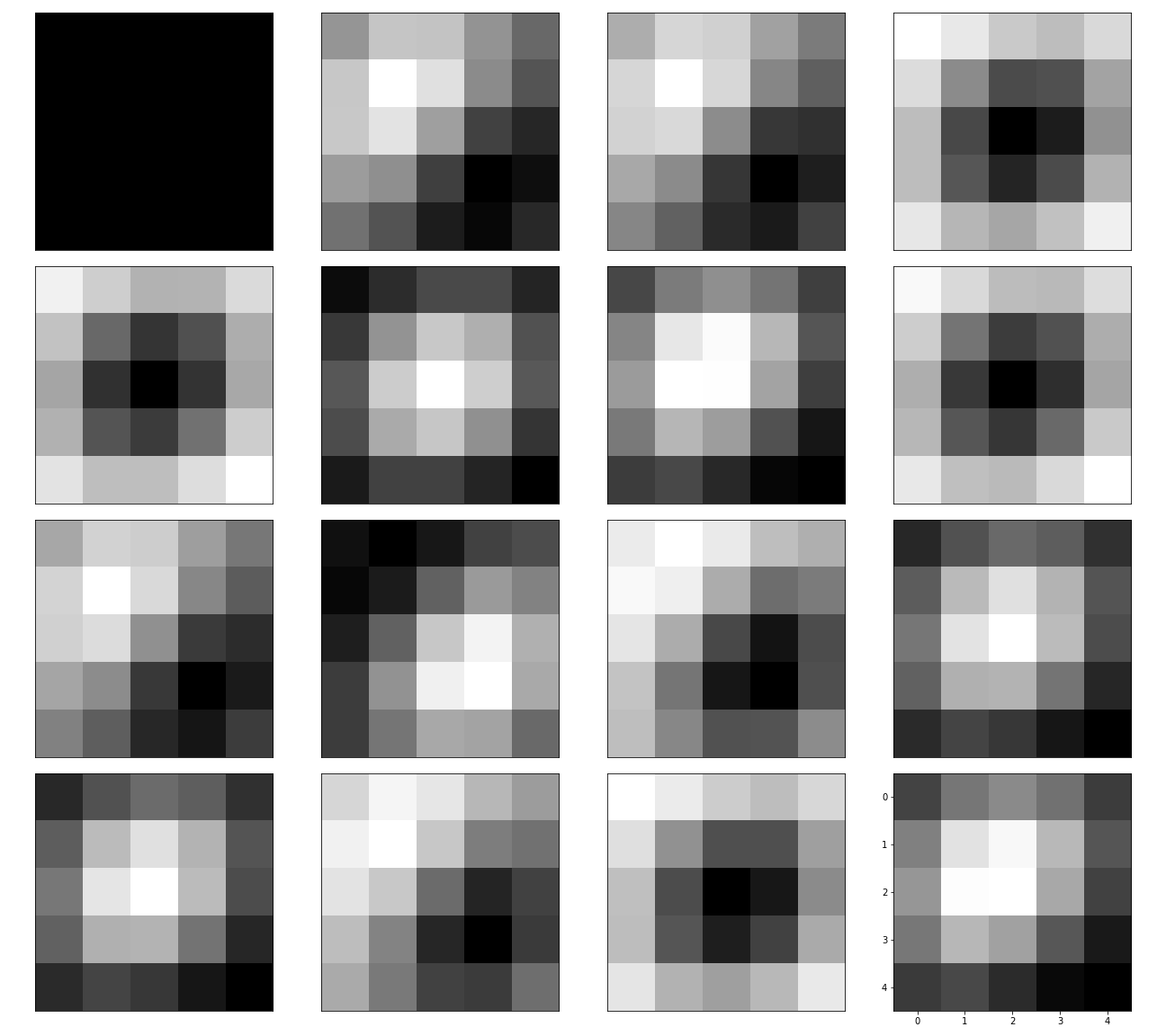}
	\caption{Scale filters $ \bA_1 $.}
	\end{subfigure}
	\caption{Scale decomposition; 16 channels; untied filters.}
	\label{fig:scale}
	\vspace{-1.5em}
\end{figure}

\begin{figure}[h!]
	\centering
	\begin{subfigure}[t]{.49\linewidth}
	\centering
	\includegraphics[width=1\columnwidth,height=0.7\columnwidth]{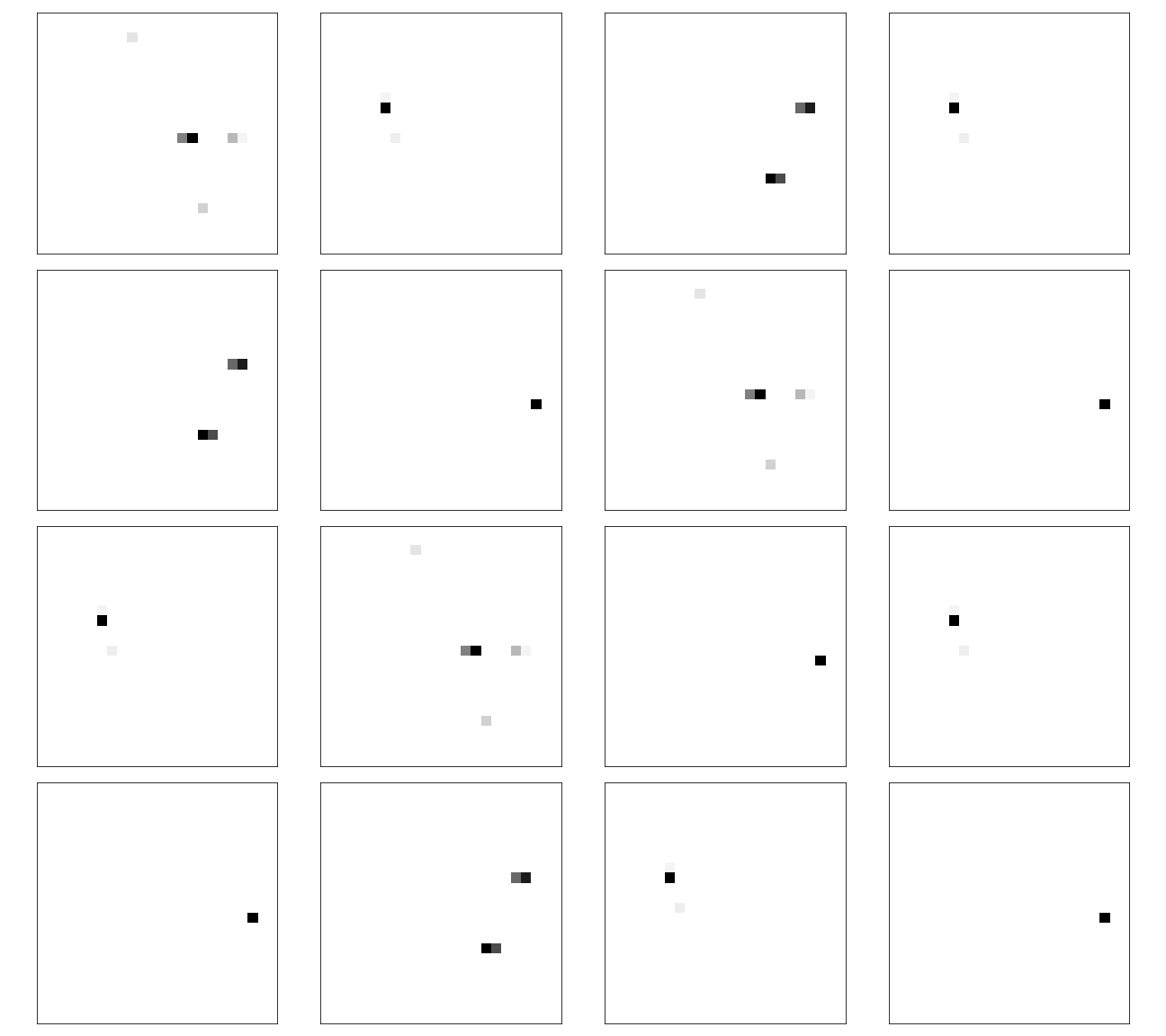}
	\caption{First layer $ \bu_1 $.}
	\end{subfigure}
	\begin{subfigure}[t]{.49\linewidth}
	\centering
	\includegraphics[width=1\columnwidth,height=0.7\columnwidth]{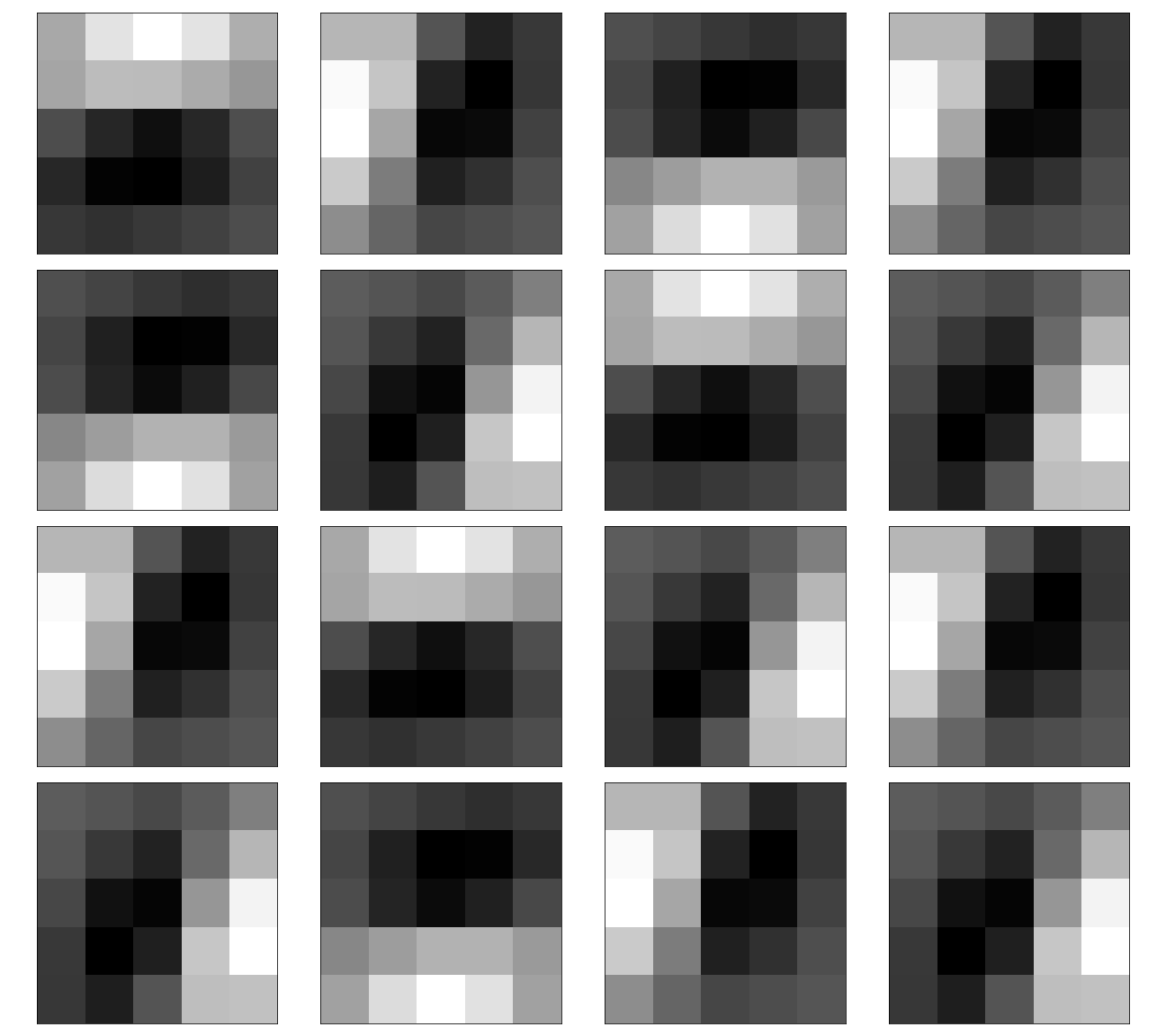}
	\caption{Detail filters $ \bB_1 $.}
	\end{subfigure}
	\caption{Detail decomposition; 16 channels; untied filters}
	\label{fig:detail}
	\vspace{-2em}
\end{figure}

\noindent\textbf{Visualization:} In \cref{fig:scale,fig:detail}, we provide a visualization of the encoded features derived by the hierarchical model and untied filters.
\cref{fig:scale} shows the scale representation $ \bx_1 $ and the corresponding learned filters $ \bA_1 $ with $ (16\times 1\times 5\times 5) $ dimensions. 
We can observe that filters learn a varied set of features, where some seem to be low frequency, but others are high frequency as well.
This visualization indicates that as the number of scale filters increases, the channels become more expressive and the representation error decreases.

\Cref{fig:scale} displays the detail signal $ \bu_1 $ and filter $ \bB_1 $ on 16 channels.
What we observe is that the encodings are very sparse, and the filters become more specialized and use more contrasting shapes.
This shows that the kind of filters that are learned for the scale and detail signals are different.



\section{Conclusion}

We proposed a generative convolutional model to analyze signals based on smooth representation (scale), and sparse contributions (detail).
This model used a recursive procedure where the scale signals were further decomposed into subsequent scale and detail components, providing higher order representations.
Such decomposition used a hierarchical structure of filters, which can be shared between layers (tied) or independent (untied).
Tied filters employed less trainable parameters and resembled the analytical process of wavelets.
We evaluated the model on a classification task on MNIST and reached 98.1\% accuracy only using 800 parameters.
Future work will further address these systems adding up/down sampling operations to obtain multiscale representations with trainable filters.


\bibliographystyle{IEEEbib}
\bibliography{refs}

\end{document}

%% file: preamble.tex
\usepackage{spconf}
\usepackage{amsmath}
\usepackage{graphicx}
\usepackage{subcaption}
\usepackage{amssymb}
\usepackage{braket}
\usepackage{bm}
\usepackage[linesnumbered,ruled,vlined]{algorithm2e}
\usepackage{hyperref}
\usepackage[nameinlink,capitalise,noabbrev]{cleveref}
\usepackage{IEEEtrantools}

\usepackage{tikz}
\usetikzlibrary{fit,calc,positioning,decorations.pathreplacing,matrix}
\usetikzlibrary{shapes,arrows}
\usetikzlibrary{calc}

\hypersetup{
    colorlinks,
    linkcolor={black},
    citecolor={black},
    urlcolor={black},
}

\def\bx{{\mathbf x}}
\def\bu{{\mathbf u}}
\def\br{{\mathbf r}}
\def\bb{{\mathbf b}}
\def\bA{{\mathbf A}}
\def\bB{{\mathbf B}}
\def\bEps{{\bm \varepsilon}}
\newcommand{\calN}{\mathcal{N}}

\DeclareMathOperator{\laplace}{Laplace}
\DeclareMathOperator{\relu}{ReLU}
\DeclareMathOperator*{\argmin}{\arg\,\min}

%% file: model.tikz
\tikzstyle{block} = [draw, fill=none, rectangle, minimum height=2em, minimum width=2em]
\tikzstyle{sum} = [draw, fill=none, circle, node distance=1cm]
\tikzstyle{cir} = [draw, fill=none, circle, line width=1mm, minimum width=0.7cm, node distance=1cm]
\tikzstyle{loss} = [draw, fill=none, color=black, ellipse, line width=0.5mm, minimum width=0.7cm, node distance=1cm]
\tikzstyle{blueloss} = [draw, fill=none, color=black, ellipse, line width=0.5mm, minimum width=2.5cm, node distance=1cm, color=black]
\tikzstyle{input} = [coordinate]
\tikzstyle{output} = [coordinate]
\tikzstyle{pinstyle} = [pin edge={to-,thin,black}]
\begin{tikzpicture}[auto, node distance=2cm,>=latex']
		cloud/.style={
			draw=red,
			thick,
			ellipse,
			fill=none,
			minimum height=1em}
		\node [input, name=x3] {};
		\node [rectangle, fill=none, node distance=1.2cm, below of=x3] (u3) {$$};

		\node [block, node distance=1cm, right of=x3] (A3) {$\bA_3$};
		\node [block, node distance=1.2cm, below of=A3] (B3) {$\bB_3$};
		\node [rectangle, fill=none, node distance=0.6cm, below of=A3] (3) {$$};
		\node [sum, node distance=0.9cm, right of=3] (sum3) {$+$};
		\node [rectangle, fill=none, node distance=1.2cm, below of=sum3] (u2) {$$};

		\node [block, node distance=1.2cm, right of=sum3] (A2) {$\bA_2$};
		\node [block, node distance=1.2cm, below of=A2] (B2) {$\bB_2$};
		\node [rectangle, fill=none, node distance=0.6cm, below of=A2] (2) {$$};
		\node [sum, node distance=0.9cm, right of=2] (sum2) {$+$};
		\node [rectangle, fill=none, node distance=1.2cm, below of=sum2] (u1) {$$};

		\node [block, node distance=1.2cm, right of=sum2] (A1) {$\bA_1$};
		\node [block, node distance=1.2cm, below of=A1] (B1) {$\bB_1$};
		\node [rectangle, fill=none, node distance=0.6cm, below of=A1] (1) {$$};
		\node [sum, node distance=0.9cm, right of=1] (sum1) {$+$};
		
		\node [output, node distance=1cm, right of=sum1] (out) {x};
		
		\draw [->] (x3) -- node[midway] {$\bx_3$} (A3);
		\draw [->] (u3) -- node[midway] {$\bu_3$} (B3);
		\draw [->] (u2) -- node[midway] {$\bu_2$} (B2);
		\draw [->] (u1) -- node[midway] {$\bu_1$} (B1);
		
        		\draw [->] (A3) -| (sum3);
		\draw [->] (B3) -| (sum3);
		\draw [->] (sum3) -- node[midway] {$\bx_2$} (A2);
		\draw [->] (sum3) -- (A2);
		
		 \draw [->] (A2) -| (sum2);
		\draw [->] (B2) -| (sum2);
		\draw [->] (sum2) -- node[midway] {$\bx_1$} (A1);

        		\draw [->] (A1) -| (sum1);
		\draw [->] (B1) -| (sum1);
		\draw [->] (sum1) -- node[midway] {$\bx_0$} (out);
		
\end{tikzpicture}